\documentclass[a4paper, 10pt, conference]{ieeeconf}      

\IEEEoverridecommandlockouts                              
\usepackage{graphicx}
\usepackage{tabularx}
\usepackage[utf8]{inputenc}
\usepackage[T1]{fontenc}
\usepackage{subcaption}

\newsavebox{\twosubbox}
\usepackage{xcolor, hyperref}
\usepackage{color,soul}
\usepackage{lipsum}
\usepackage{hyperref}
\usepackage{array, multirow, bigdelim, makecell, booktabs} 
\title{\LARGE \bf
Instructing Hierarchical Tasks to Robots by Verbal Commands
}

\author{Péter Telkes$^{1}$, Alexandre Angleraud$^{1}$ and Roel Pieters$^{1}$
\thanks{$^{1}$Cognitive Robotics group, Unit of Automation Technology and Mechanical Engineering, Tampere University, 33720, Tampere, Finland; {\tt\small firstname.surname@tuni.fi}}%
}

\begin{document}

\maketitle
\thispagestyle{empty}
\pagestyle{empty}

\begin{abstract}
Natural language is an effective tool for communication, as information can be expressed in different ways and at different levels of complexity. Verbal commands, utilized for instructing robot tasks, can therefor replace traditional robot programming techniques, and provide a more expressive means to assign actions and enable collaboration. However, the challenge of utilizing speech for robot programming is how actions and targets can be grounded to physical entities in the world. In addition, to be time-efficient, a balance needs to be found between fine- and course-grained commands and natural language phrases. 
In this work we provide a framework for instructing tasks to robots by verbal commands. The framework includes functionalities for single commands to actions and targets, as well as longer-term sequences of actions, thereby  providing a hierarchical structure to the robot tasks.
Experimental evaluation demonstrates the functionalities of the framework by human collaboration with a robot in different tasks, with different levels of complexity. The tools are provided open-source at  \textit{\small \url{https://petim44.github.io/voice-jogger/}}

\end{abstract}

\section{Introduction}\label{sec:intro}
Speech and natural language is the most common modality for interaction between humans, due to its expressive nature and its ability for dialogue. Communication between humans and robots can benefit as well from such functionalities, and, as a research field, efforts to achieve this have been ongoing since several decades \cite{Tellex2020robots}.
The main challenge in instructing robots by verbal commands is how natural language can be easily grounded to the actions of a robot. On the one hand this can be done by directly matching input commands to a single robot action or target location in the world (i.e., lexical grounding), thereby effectively performing tele-operation. On the other hand, learning-based approaches can generalize input commands to desired outcomes in the world, from Large Language Models (LLM), possibly assisted by visual perception (vision-language pre-training, or VLP).

While lexical grounding provides a one-to-one connection from commands to actions, learning-based models do not provide this, making them less reliable to achieve desired outcomes. 
\begin{figure}[t]
  \centering
  \includegraphics[width=0.48\textwidth]{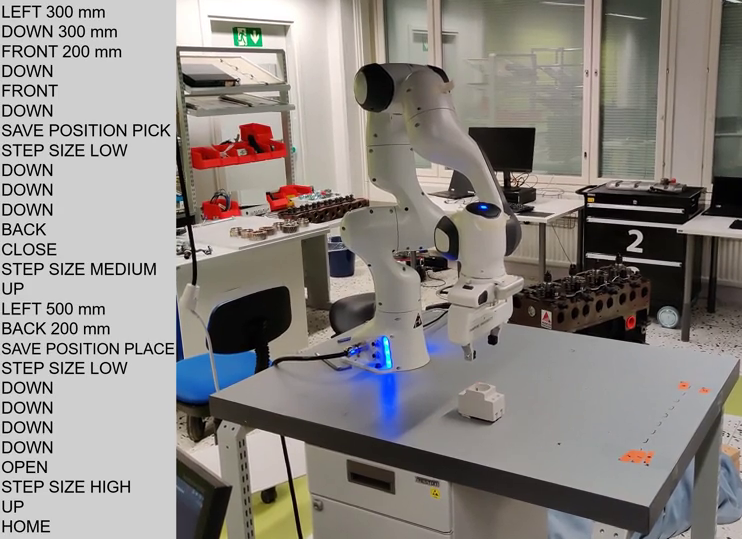}
  \caption{Instructing robots by speech requires fine-grained verbal commands (left) that directly refer to actions and targets in the workspace. This work presents methods that extend verbal command instructions to hierarchical tasks.}
    \label{fig:overall}
\end{figure}
In both cases basic robot functionalities are needed that act in the world, such as movement primitives in world space, gripper actions and higher-level hierarchical tasks that execute multiple actions in sequence. Most importantly, the grounding of (verbal) natural language commands to robot actions requires manual selection of what robot actions are needed for what tasks. While LLMs and VLP can help in extracting relevant information from language and the scene, they cannot alone solve this grounding problem.

In this work, we present a framework for verbal commands as tool to instruct tasks to robots, including automatic speech recognition and the grounding of commands to robot actions and targets in the work space (see Fig. \ref{fig:overall}). Single and multiple command phrases from natural language enable different functionalities to be included such as task hierarchies and task concatenation, without explicitly relying on high-level task planners. These capabilities are of particular importance to ensure smooth and efficient collaboration between human and robot, as continuously using fine-grained language commands is time-consuming (see Fig. \ref{fig:overall}). Course-grained commands enable high-level functionalities and long-term sequences of actions as demonstrated by our work with two hierarchical tasks for gear assembly. Our developments and tools are integrated in a robotic system and numerous experiments with single commands and long-term tasks demonstrate the results. %




The contributions of this work are as follow:
\begin{itemize}
    \item Framework for instructing robots by verbal commands with basic robot functionalities
    \item Grounding of commands to robot actions and targets in the scene
    \item High-level functionalities to support task hierarchies and concatenation
    \item Experimental demonstration of the framework with representative tasks
\end{itemize}


\section{Related Work}\label{sec:related_work}
\subsection{Human-Robot Collaboration}
Collaboration between human and robot can be supported in many ways by different technologies \cite{villani2018survey}. While physical interaction (e.g., hand-guiding or learning by demonstration) and visual perception (e.g., for object and gesture detection) are popular modalities \cite{gross2023communicative}, also audio perception in the form of natural language and speech is common \cite{Tellex2020robots}. Success of the collaboration depends on different factors, such as technical quality of the tools, and how communication between human and robot is enabled \cite{marge2022spoken}. 
In context to this work, the collaboration between human and robot has the purpose to provide robot instructions via speech. This means that communication is only directed from human to robot, and human verbal commands (i.e., natural language) need to be recognized and interpreted to be utilized for robot task execution. 
Several earlier works have developed capabilities that enable robots to act on voice commands \cite{Tellex2020robots}, with different levels of complexity and even with additional sensor modalities to support the understanding of the world and tasks within it. 

The related work therefore covers the state of the art in natural language processing and further divides robotics approaches in two directions; lexically-grounded methods and learning-based methods (see Table \ref{tab:commands_vs_learning}). 

\subsection{Natural language processing}
Speech recognition and natural language processing (NLP) are two related fields, which both have been propelled forward recently due to the application of deep learning models \cite{yu2016automatic, Otter2021surveyDLNLP}. For this work, we focus on NLP for the understanding of text and assume that a correct recognition of speech (by e.g. Vosk \cite{alphacephei:vosk}) provides natural language (words or sentences) in textual format. The interpretation of language by Large Language Models (LLM), such as BERT \cite{devlin2018bert} and PaLM \cite{chowdhery2022palm}, 
have demonstrated capabilities including language understanding and generation, reasoning, and code-related tasks. As such, to enable robot instructions, they still need the explicit grounding of verbal commands to actions of the robot or objects in the scene. 


\subsection{Lexical-grounding of robot commands}
Commanding robots from natural language by lexical grounding has been demonstrated for single verbal commands \cite{stenmark2014describing}, command pairs \cite{Angleraud2021} and short phrases \cite{tenorio2010teaching}. In most of these cases, words to be recognized and tasks to executed are predefined, limiting the applicability of the methods to new situations and new tasks.
Verbal commands have also been utilized to aid in hand-guided task programming \cite{ionescu2021programming}, thereby effectively enabling multi-modal robot programming. A list of supported single word commands can be commanded and, separately, different parameters can be set via an app to adapt different parameters for robot motion and skills.
Speech interfaces have also been utilized for modification of robot skills \cite{frasca2021enabling}, utilizing bi-directional and high-level communication between human and robot. Interaction scenarios are developed that show long human commands and instructions to modify robot knowledge.  
In \cite{suddrey2016teaching} the learning of complex hierarchical tasks from natural language instruction is demonstrated, utilizing hierarchical task networks (HTNs) and queries from robot to human in case tasks are unknown. Similarly, in \cite{suddrey2022learning} behavior trees are utilized to learn and perform novel complex tasks. As such, both these approaches focused on the planning of complex tasks and the learning of task hierarchies and behaviors. 
Complex natural language instructions have also been utilized for planning in linear temporal logic formulation \cite{boteanu2016model}, with incomplete world knowledge \cite{nyga2018grounding} and for the ordering \cite{pramanick2020decomplex} and allocation \cite{behrens2019specifying} of tasks to be executed.

\begin{table}[t]
\centering
\caption{Comparison between two approaches of utilizing speech in robotics}
\label{tab:commands_vs_learning}
\begin{tabular}{p{0.07\textwidth}|p{0.18\textwidth}|p{0.16\textwidth}}

&\textbf{Lexical grounding}       & \textbf{Learning-based}              \\ \hline\hline
Approach & Match words directly to robot actions and targets in the workspace & Use high-level natural language instructions to learn robot actions in the workspace \\ \hline
Commands & Fine-grained & Course-grained \\ \hline
Vocabulary & Limited & Expressive \\ \hline
Actions & Short-time & Mid- to long-term \\ \hline
Response & Fast & Slow \\ \hline
References & Single commands \cite{tenorio2010teaching, ionescu2021programming} & CLIPort \cite{Shridhar2022cliport} \\ 
& Command pairs \cite{Angleraud2021} & PaLM-E \cite{driess2023palm}\\
& Command sentences \cite{stenmark2014describing} & Interactive language \cite{lynch2022interactive} \\
& Hierarchical tasks \cite{suddrey2016teaching, suddrey2022learning}& \\ 
& This work & \\ \hline
\end{tabular}
\end{table}

What all of the mentioned approaches have in common is that verbal commands directly result in robot actions, or indirectly through task plans. As hand-crafted algorithms extract relevant words from input, this means approaches are typically limited to a predefined vocabularies and action plans.
 

\subsection{Learning-based robot commands}
Natural language as input for robot tasks has seen recent interest by generating task policies that are language-conditioned \cite{luketina2019survey}. Even more recent are the inclusion of other modalities, in addition to natural language, to establish the link between words and perception \cite{shao2020concept, Shridhar2022cliport, sharma2022correcting}, for tasks such as robot object manipulation and motion trajectory generation. While these methods prove well-capable of transforming natural commands to robot actions, the question remains whether such vision-and-language pretraining improves lexical grounding \cite{yun2021does}. This embodiment issue is addressed in \cite{driess2023palm} and \cite{ahn2022doasi} which directly incorporate real-world continuous sensor modalities into a LLM (PaLM) to plan and execute long horizon tasks, i.e., robot planning on table-top and object manipulation and navigation in office kitchen environments, respectively. Another example is embodied BERT (emBERT) \cite{suglia2021embodied}, which grounds language instructions against visual observations and actions to take in an environment, with the aim of home and office navigation. Finally, real-time (i.e., 5Hz) interaction by speech is presented in \cite{lynch2022interactive}, which demonstrates the instruction of multiple robots on table-top setting at the same time.




For both lexical grounding and learning-based approaches, still a connection needs to be made to robot actions and the speech commands. While this connection can be learned, as demonstrated by learning-based approaches, only a person can verify whether the learned grounding is correct. In case of high-performance or safety critical tasks (e.g., industry or search-and-rescue, as compared to household actions by a service robot) such uncertainty might not be acceptable. Our lexical-grounding approach is human-coordinated and aims to enable basic and hierarchical tasks to be commanded from human to robot, with both fine and course-grained commands.










\section{Approach}\label{sec:system}

Our approach divides robot instructions into basic and hierarchical commands (see Table \ref{tab:commands}), enabling both low- and high-level commands to be instructed at the same time. New target poses and tasks can be defined at any time with any combination of basic and hierarchical commands.

\subsection{Grounding of commands}
Grounding of verbal commands is done by connecting individual words to actions or targets, when they are detected from automatic speech recognition. Robot actions can be named with any spoken word and multiple words can refer to the same action. This similarly applies to robot pose targets (command \texttt{save}) and new defined (hierarchical) tasks (command \texttt{record}). Verbal commands are also grounded to robot motion in case of single direction motion (command \texttt{<direction> <value>}) and other robot actions (\texttt{home}, \texttt{open}, \texttt{close}, \texttt{rotate}, etc.).



\subsection{Basic commands} 
All relevant commands needed to enable a functional robot are referred to as basic commands. This includes commands to start and stop the robot system and individual motion actions of the different components (single direction step motion, continuous point-to-point motion, gripper motion, etc.).
Depending on the functionality of the command, different instruction combinations can be given as arguments, as described in Table \ref{tab:commands}. Basic commands can be utilized at any point during robot operation and can be used for hierarchical tasks in short- or long-term form.


\subsection{Hierarchical task commands}
Hierarchical tasks offer the ordering of robot actions in different ways (see Table \ref{tab:commands}). For example, task recording (initiated by command \texttt{record}, terminated by command \texttt{finish}) enables task hierarchies to be assigned by the human operator. In its simplest form these contain a list of actions and commands that are executed in sequence, as described by few included template commands (i.e., \texttt{pick, place, stack, push}). 
Task concatenation is enabled by the \texttt{and} command, by appending following commands to a list. Combining multiple robot motion actions is enabled by the \texttt{then} command, which computes the spoken motion phrases to one single motion command outcome. For example, multiple directional commands in sequence will simply be added up as a robot pose target (i.e., \texttt{down then down then left 200 mm} will result in a single robot pose translation target with the spoken offsets).
Task repetition is enabled by the commands \texttt{again} and \texttt{repeat}, given suitable input for executing the motion commands (i.e., list of robot poses and number of repetitions).

\begin{table}[!t]
\centering
\caption{List of available basic and hierarchical commands for instructing robot motion and tasks. 
}
\label{tab:commands}
\scriptsize
\begin{tabular}{l|p{0.23\textwidth}}
\hline
\textbf{Basic commands}       & \textbf{Explanation}              \\ \hline
\texttt{start robot}            & Initialize robot                                        \\
\texttt{stop robot}             & Shut-down robot                                          \\
\texttt{set mode step}          & Set robot to move in steps                   \\
\texttt{set mode continuous}    & Set robot to move continuous                         \\
\texttt{save position <string>} & Save robot pose as \texttt{<string>} \\
\texttt{position <string>  }    & Move robot to pose \texttt{<string>}       \\
\texttt{<direction> <value>}   & Move robot \texttt{<value>} \texttt{\{up, down, left, right, back, front\}}                    \\
\texttt{stop execution}         & Stop robot motion                                     \\
\texttt{step size <value>}      & Set robot motion step size               \\
\texttt{open}              & Open gripper                                          \\
\texttt{close}             & Close gripper                                         \\
\texttt{rotate <value> }   & Rotate gripper by \texttt{<value>} deg.                   \\
\texttt{home}                   & Move robot to home position                           \\ \hline
\textbf{Hierarchical commands}       & \textbf{Explanation}\\ \hline
\texttt{record <string>}        & Start recording commands as \texttt{<string>}\\
\texttt{finish}                 & Stop recording command sequence \\
\texttt{task <string>}          & Execute command sequence \texttt{<string>}\\ 
\texttt{repeat <task> <string> }    & Repetition of \texttt{<task>}, continuously or specified number of times by \texttt{<string>}\\
\texttt{repeat <task> <list>  }    & Repetition of \texttt{<task>}, until \texttt{<list>} of positions is empty\\
\texttt{pick <string> }    & pick object from pose \texttt{<string>} \\     
\texttt{place <string> }    & place object to pose \texttt{<string>} \\
\texttt{stack <string> <value>}    & Pick and place on pose \texttt{<string>}  with vertical offset \texttt{<value>}\\
\texttt{push <string> <direction>}    & Planar motion at pose \texttt{<string>} starting from \texttt{<direction>}\\ \hline

\hline
\end{tabular}
\end{table}



New (hierarchical) commands can be easily defined, utilizing existing commands and by providing desired arguments. Several typical robot manipulation commands, such as pick-and-place, wiping and polishing have been developed this way and are demonstrated individually and as integrated in two assembly use cases in Section \ref{sec:results}.

\subsection{Robot Control Architecture}
The architecture for controlling the robot by verbal commands includes modules for command detection, robot motion planning and robot control as follows. Command detection recognizes speech commands, adds these to a queue and publishes them as suitable messages over ROS topics. This means that the system is continuously listening to audio and adds new robot tasks when they are received. A high priority topic is used for the \texttt{stop} command, which allows to bypass the queue and executes a stopping motion immediately, with suitable motion profile. In case of teaching positions or task hierarchies, the system stores them locally in custom XML format with commanded name, to be used when requested. The motion planning module relies on MoveIt2 and is continuously waiting for new motion commands from the command topics. Finally, robot control utilizes the Franka Control Interface (FCI) to achieve real-time execution of control actions.

\section{Results and Discussion}\label{sec:results}

\subsection{Hardware and implementation}
The experimental setup includes a Franka Emika robot with parallel gripper. 
Robot motion control utilizes Moveit2 motion planner for generating work space motion and ROS (ROS.org) handles all communication.
Speech of a person is recorded by a standard microphone (16kHz audio signal), transmitted to the main PC and passed through a voice activity detector \cite{SileroVAD}, to identify if the individual audio packet has human voice or not. A time-delayed filter is applied on the running stream of audio packets to consider the natural pause in human speech while speaking. For speech recognition, Vosk \cite{alphacephei:vosk} is utilized, which is trained on The People's Speech dataset \cite{galvez2021people}. The model is configured by filtering out unnecessary words, which are unsuitable for robot instructions, and by including words and sentence instructions that reflect the content of speech to be expected. All packages, tools and videos are open-source available to replicate our work at \textit{\small \url{https://petim44.github.io/voice-jogger/}}.

\subsection{Speech recognition}
The accuracy of speech recognition was tested in a clean and noisy environment with a non-native English speaker (see Table \ref{tab:speech_recog_experiment}). In the clean case, verbal commands are recognized without any failures. In the noisy environment the audio contains a background noise similar to a busy restaurant with other people speaking, which deteriorates the detection rate slightly. Speech recognition is also evaluated in the use cases for single and hierarchical tasks, which resulted in a high success rate (see Table \ref{tab:results_speech_tasks}). 

\begin{table}[ht]
  \scriptsize
  \begin{center}
    \caption{Speech recognition experiment results}
    \label{tab:speech_recog_experiment}
    \begin{tabular}{l | c | c }
      \hline
    \textbf{Command} & \textbf{Clean} & \textbf{Noisy}\\
      \hline 
    \texttt{start robot}	& \textbf{10/10} & \textbf{10/10} \\
    \texttt{stop robot}	& \textbf{10/10} & \textbf{10/10} \\
    \texttt{move up} 	& \textbf{10/10} & \textbf{10/10}  \\
    \texttt{move down}  & \textbf{10/10} & \textbf{10/10}  \\
    \texttt{move left} 	& \textbf{10/10} & \textbf{10/10}  \\
    \texttt{move right} & \textbf{10/10} & \textbf{08/10}  \\
    \texttt{move back}	& \textbf{10/10} & \textbf{10/10} \\
    \texttt{move front} & \textbf{10/10} & \textbf{09/10}  \\
    \texttt{stop execution} & \textbf{10/10} & \textbf{10/10} \\
    \texttt{step size <value>} & \textbf{10/10} & \textbf{10/10}  \\
    \texttt{open tool} & \textbf{10/10} & \textbf{07/10}  \\
    \texttt{close tool} & \textbf{10/10} & \textbf{07/10}  \\
    \texttt{rotate tool <value>} & \textbf{10/10} & \textbf{10/10} \\
    \texttt{save position <value>} & \textbf{10/10} & \textbf{10/10}  \\
    \texttt{load position <value>} & \textbf{10/10} & \textbf{10/10}  \\
    \texttt{home} & \textbf{10/10} & \textbf{08/10}  \\
      \hline
    \end{tabular}
  \end{center}
\end{table}

\subsection{Single tasks}
Single tasks refer to short robot actions, which by definition, can have internal hierarchical structure. This is demonstrated by a pick and place action as it is composed into hierarchical tasks by verbal instructions (see Table \ref{tab:pickplace}). From the basic and hierarchical commands available, first robot end-effector poses are defined (\texttt{save position <value>}) in combination with intermediate robot motion and gripper actions (e.g., \texttt{move, close, open}), see left column in Table \ref{tab:commands}. As a result, these robot poses are utilized for motion commands and included in a recorded hierarchical task, see right column in Table \ref{tab:commands}. This effectively means that from an initial low-level set of commands (i.e., 27 verbal commands), human instructions can be narrowed down to 19 commands and, finally, a single task command, to achieve the same robot action outcome.

Table \ref{tab:results_speech_tasks} describes the results of several single tasks that are commanded by verbal instructions. During the preparation phase specific robot poses and motion sequences are defined as required for the tasks, in few single and multiple commands. During the execution phase robot actions utilize these poses and motions for specific objects and target applications. Repetition of single tasks towards automated applications is possible as well, by specifying the task and how often it should be repeated. In case suitable, offsets in motion commands can be included to achieve more complex operations, such as wiping and polishing a larger surface.
Fig. \ref{fig:single_tasks} depicts several snapshots of the single tasks.

\begin{table}[t]
\centering
\caption{A pick and place task composed into hierarchical tasks by verbal instructions. The original query required 27 commands and, with tasks composed to replace step-wise instructions, the query is narrowed down to 19 commands and, finally, a single task command.\label{tab:pickplace} }
\scriptsize
\begin{tabular}{l l | l l }
\hline
\textbf{Commands} & \textbf{Task} & \textbf{Commands} & \textbf{Task} \\
\hline
\texttt{left 300 mm} & \hspace{-0.5em}\rdelim\}{5}{*}[\texttt{\textbf{pick}}]  & \texttt{position} \texttt{\textbf{pick}} & \hspace{-0.5em}\rdelim\}{15}{*}[\texttt{\textbf{one}}]  \\
\texttt{down 300 mm }&  & \texttt{step size low} &  \\
\texttt{down} & & \texttt{record} \texttt{\textbf{one}} \\
\texttt{front} & & \texttt{down} ($\times 2$) \\
\texttt{down} &  & \texttt{back} \\
\texttt{save position} \texttt{\textbf{pick}} & & \texttt{down} \\
\texttt{step size low} & & \texttt{close} \\
\texttt{down} ($\times 3$)& & \texttt{step size medium}\\
\texttt{back} & & \texttt{up}\\
\texttt{close} & & \texttt{Position} \texttt{\textbf{place}}\\
\texttt{step size medium} &  \hspace{-0.5em}\rdelim\}{3}{*}[\texttt{\textbf{place}}] & \texttt{step size low}\\
\texttt{up} & & \texttt{down} ($\times 3$)\\
\texttt{left 500 mm} & & \texttt{open}\\
\texttt{back 200 mm} & & \texttt{position} \texttt{\textbf{place}}\\
\texttt{save position} \texttt{\textbf{place}} & & \texttt{home}\\
\texttt{step size low} & & \texttt{finish}\\
\texttt{down} ($\times 4$)& & \\
\texttt{open} & & \\
\texttt{step size high} & & \\
\texttt{up} & & \\
\texttt{home} & & \\
\hline
\end{tabular}
\end{table}

\subsection{Hierarchical tasks}
To demonstrate further the capabilities of verbal command instructions, experiments were also performed with longer-term tasks with complex assembly steps. This included the assembly of a helical gear system with six assembly steps and a planetary gear system with nine assembly steps. In both cases human verbal instructions guided the assembly procedure and, in the case of the planetary gear, also contained three manual and collaborative tasks. Similar to single tasks, a preparation phase records specific robot poses and motion sequences, by hand-guiding and verbal commands. The execution phase entails the actual assembly task, coordinated by the person with verbal commands. Table \ref{tab:results_speech_tasks} describes the results of these hierarchical tasks averaged over ten trials each. 

\subsubsection*{Helical gear assembly} The helical gear consists of four parts (two gears, front and back plate) and requires six steps for assembly, executed by the robot. This includes five pick-and-place actions with different locations and one pushing action to connect the two gears together. Fig. \ref{fig:planetary} depicts several snapshots of the helical gear assembly task.

\subsubsection*{Planetary gear assembly} 
\begin{figure*}[htbp]
\subcaptionbox{\label{fig:wipe_1}}{%
  \includegraphics[height=0.205\linewidth]{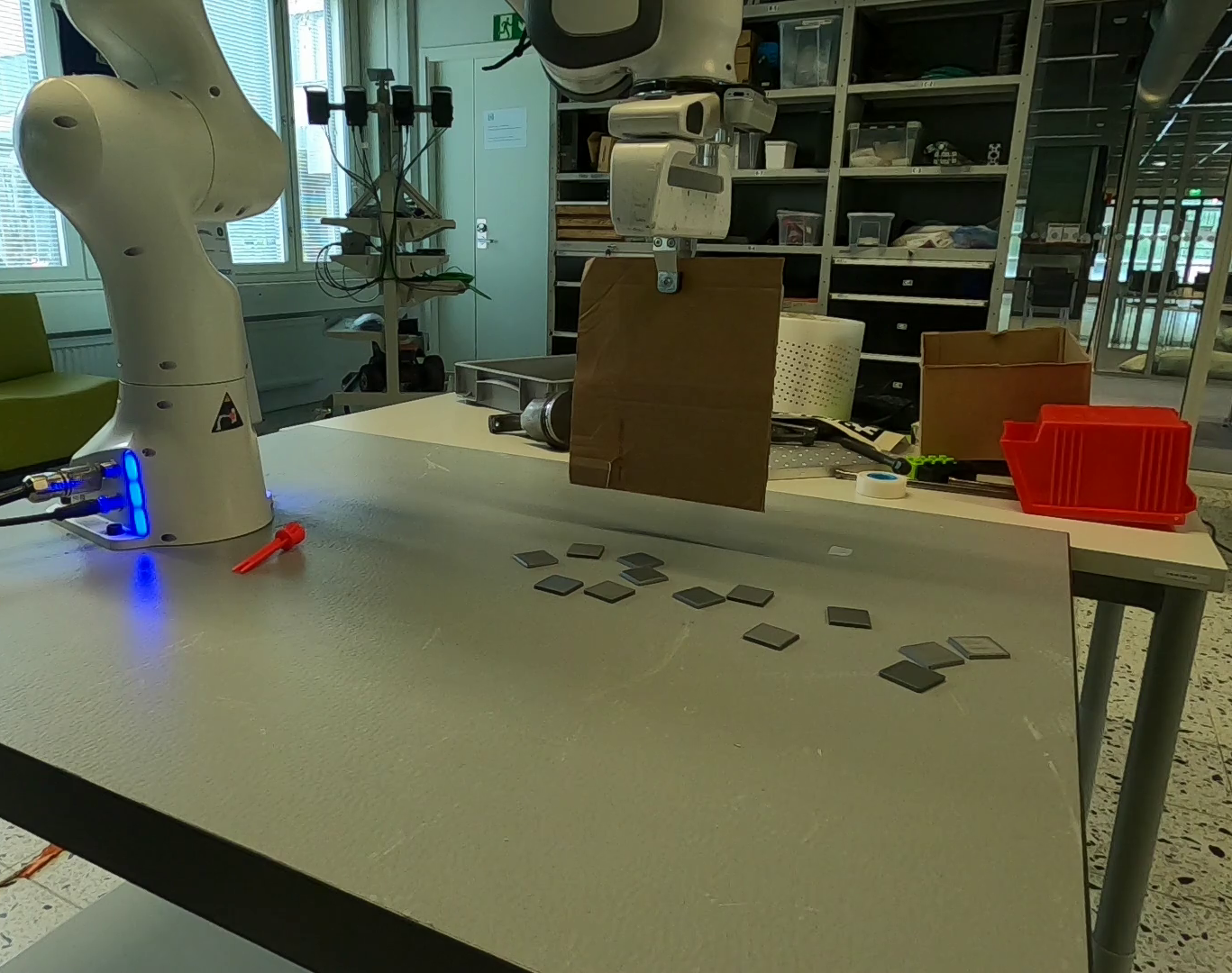} 
}\hfill
\subcaptionbox{\label{fig:wipe_2}}{%
  \includegraphics[height=0.205\linewidth]{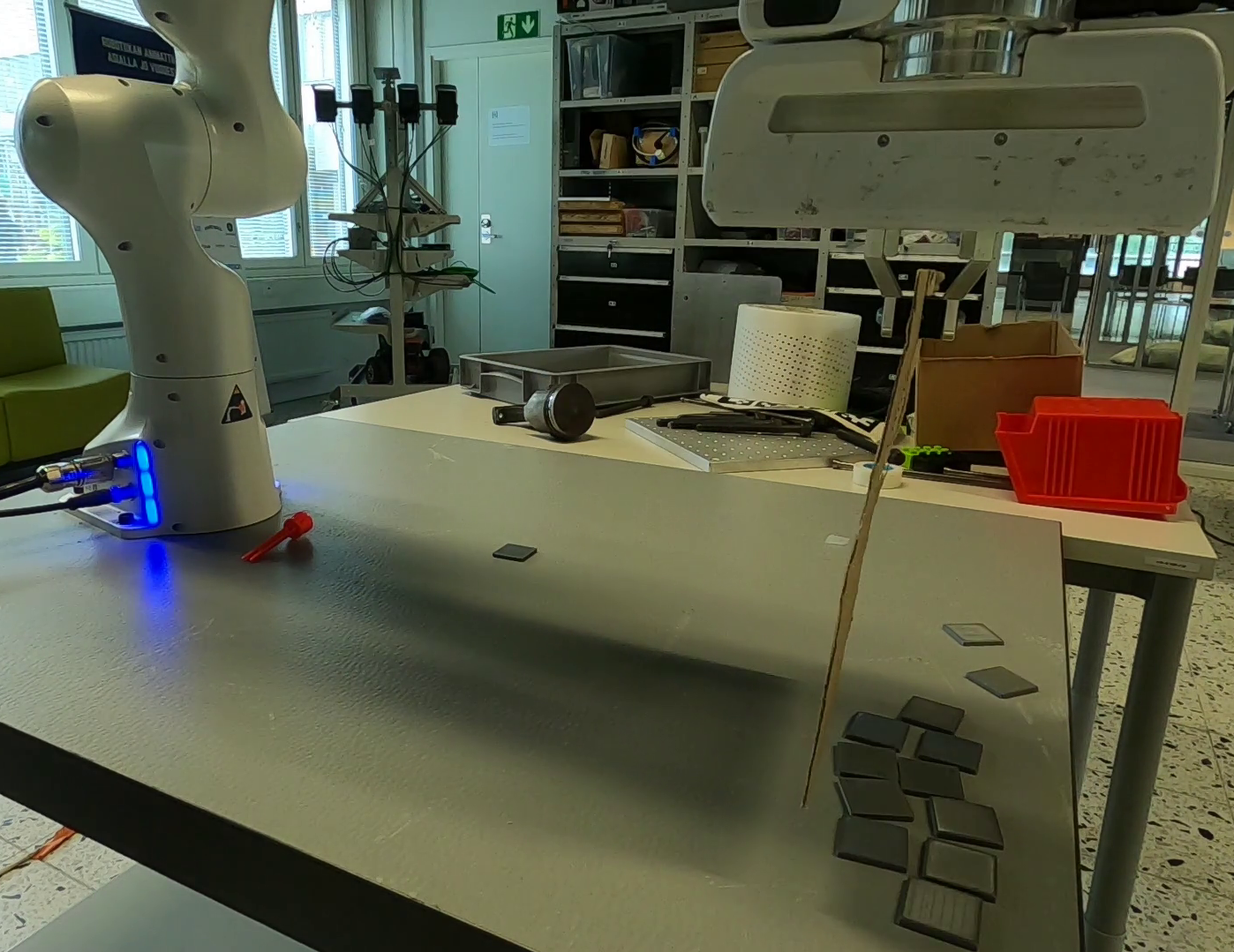} 
}\hfill
\subcaptionbox{\label{fig:wipe_3}}{%
  \includegraphics[height=0.205\linewidth]{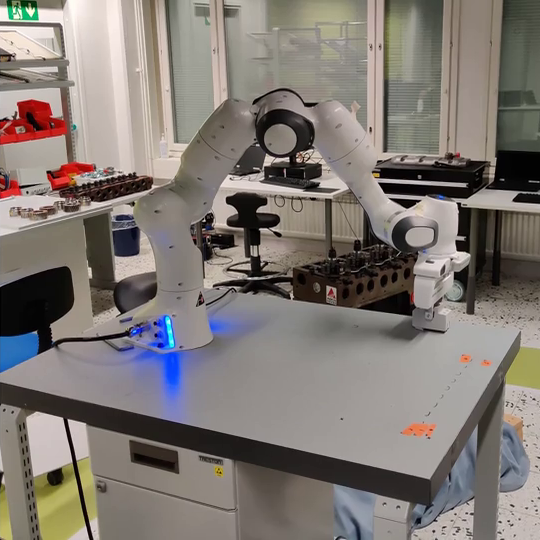} 
} \hfill
\subcaptionbox{\label{fig:wipe_4}}{%
  \includegraphics[height=0.205\linewidth]{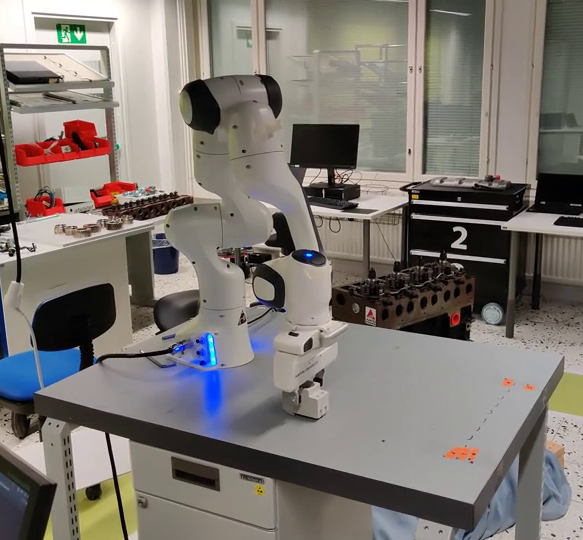} 
}  \caption{Snapshots of verbally commanded single tasks: before (a) and after (b) the wiping task, picking (c) and placing (d). \label{fig:single_tasks}}

\subcaptionbox{\label{fig:helical_1}}{%
  \includegraphics[height=0.18\linewidth]{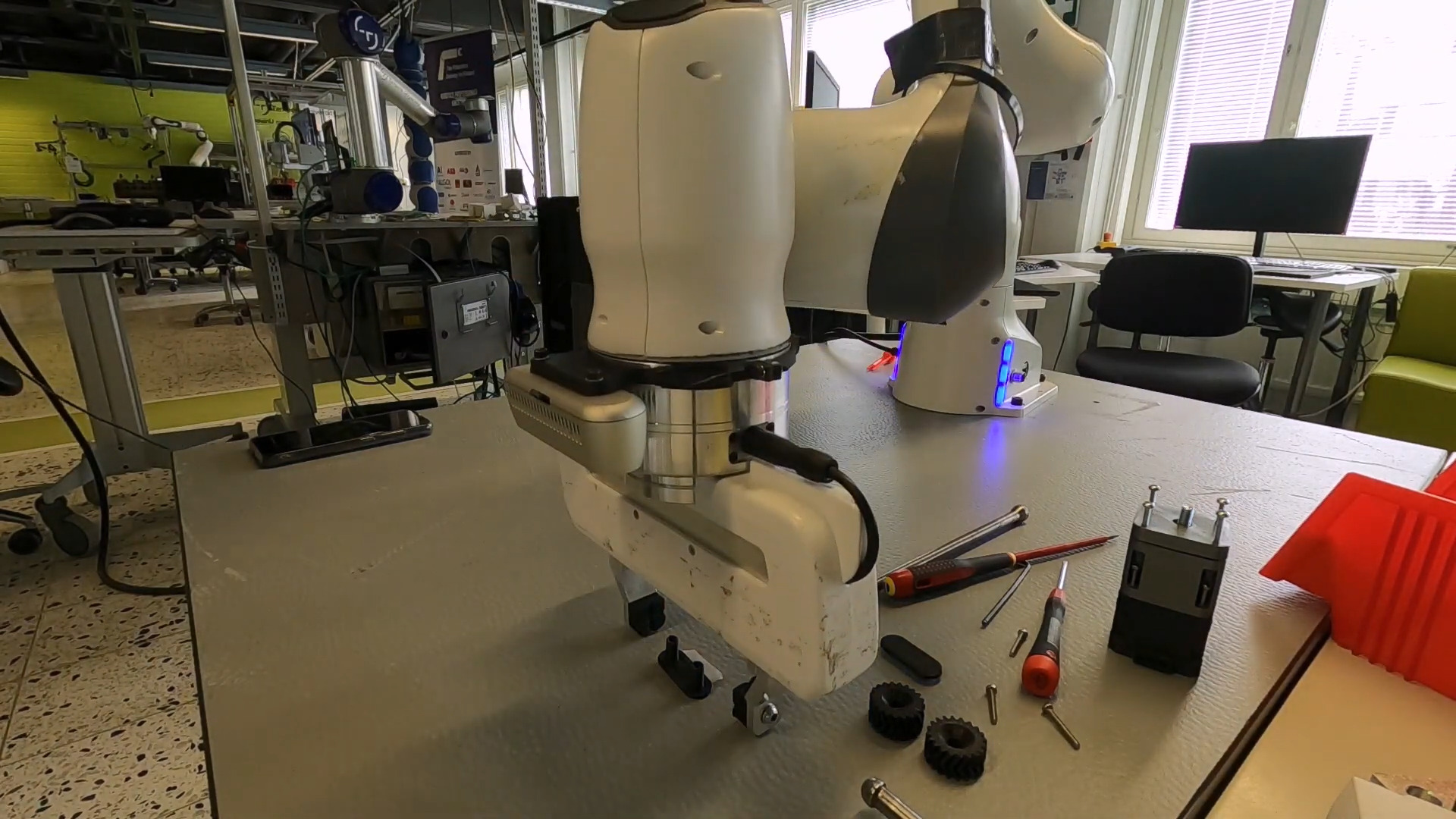} 
}\hfill
\subcaptionbox{\label{fig:Helical_2}}{%
  \includegraphics[height=0.18\linewidth]{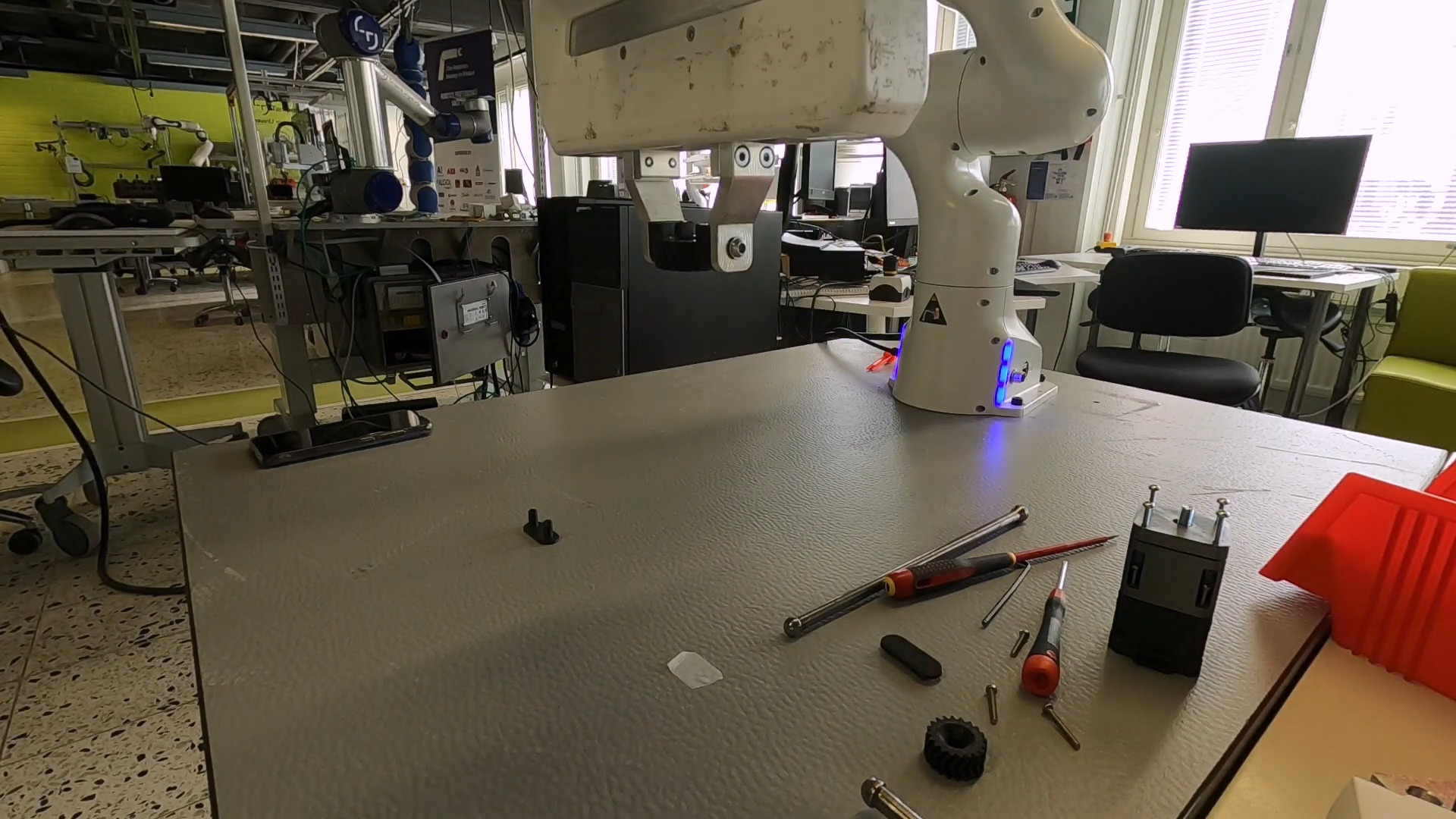} 
}
\subcaptionbox{\label{fig:Helical_3}}{%
  \includegraphics[height=0.18\linewidth]{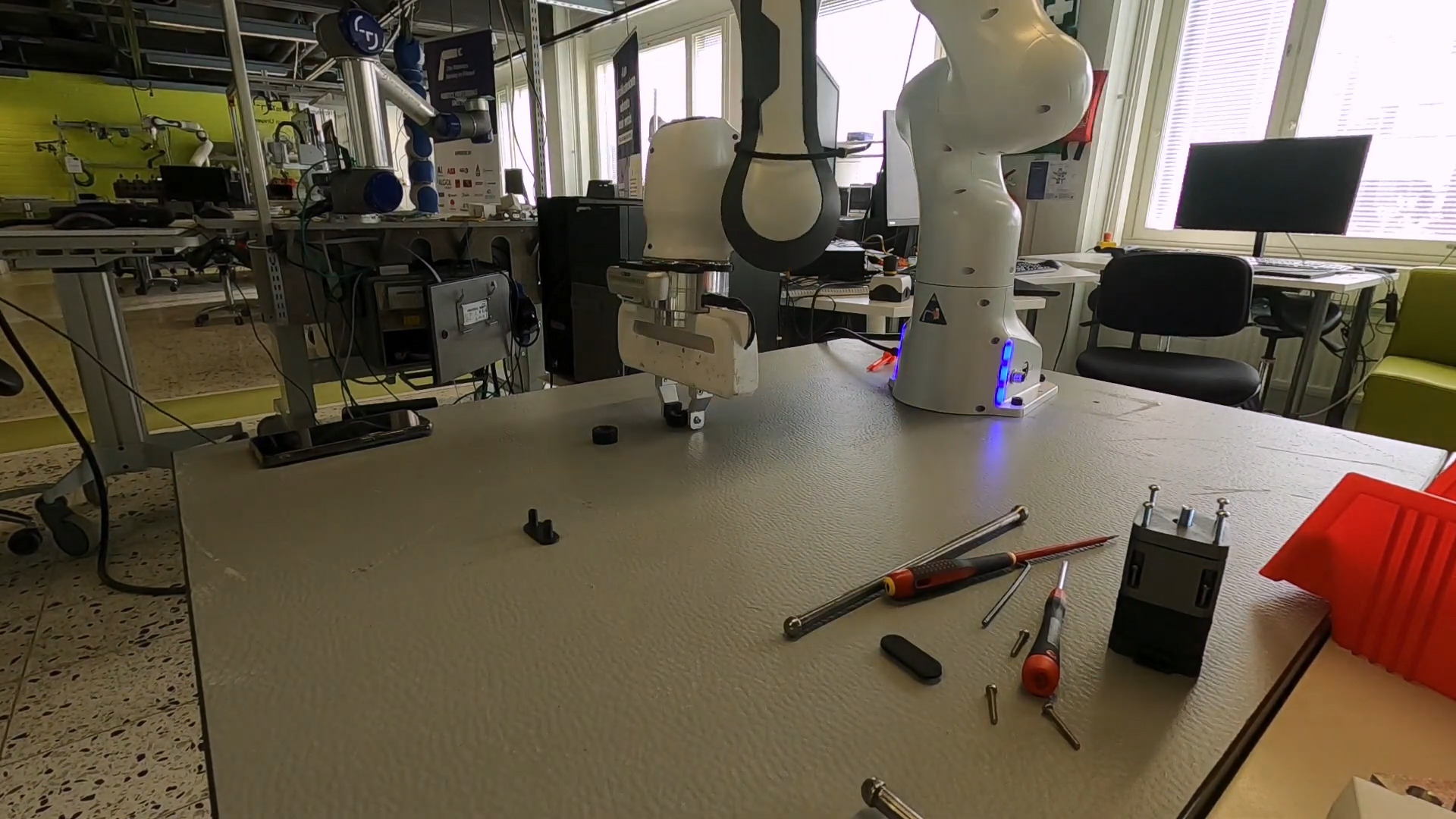} 
}
\subcaptionbox{\label{fig:helical_4}}{%
  \includegraphics[height=0.18\linewidth]{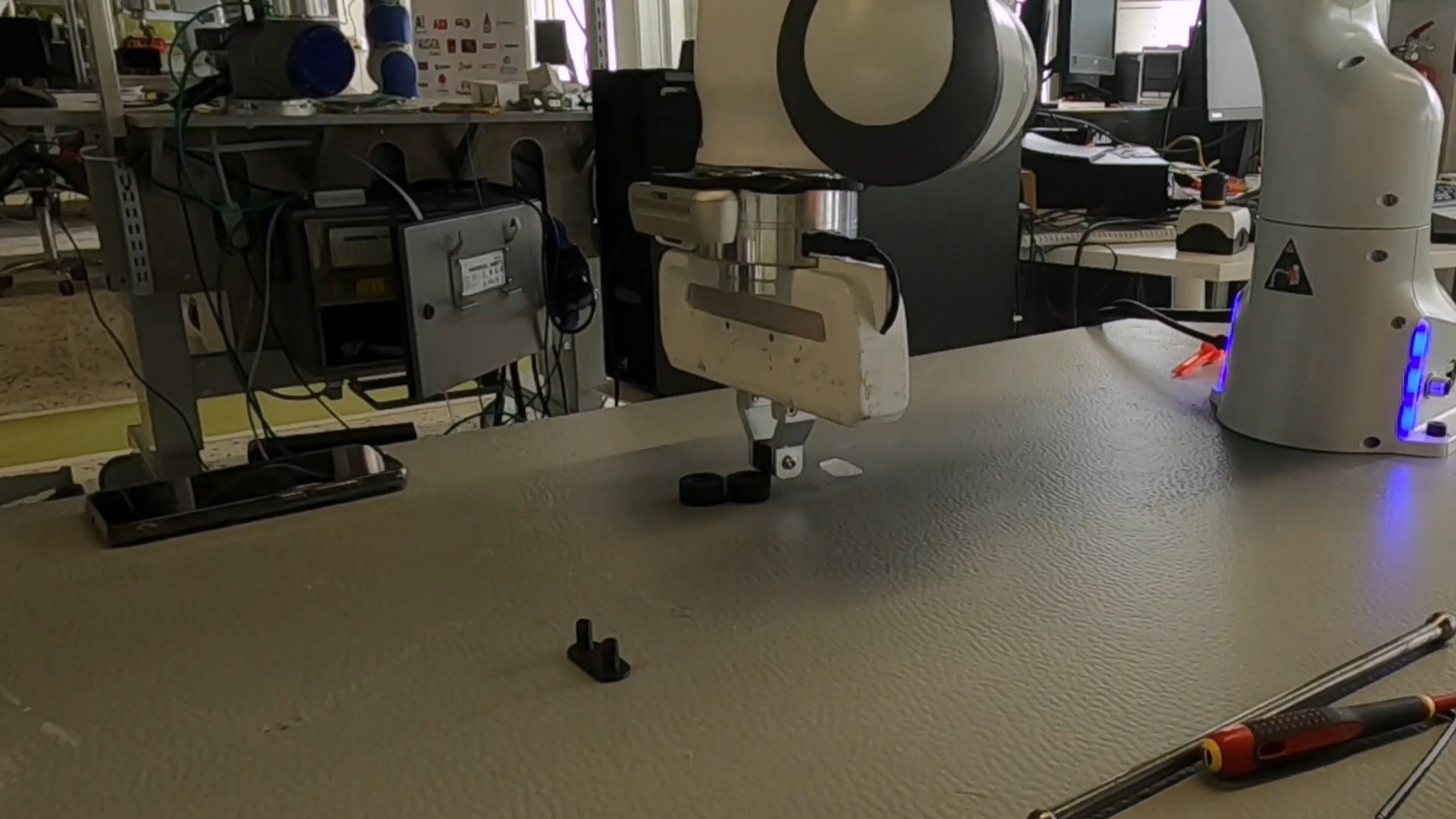} 
}\hfill
\subcaptionbox{\label{fig:Helical_5}}{%
  \includegraphics[height=0.18\linewidth]{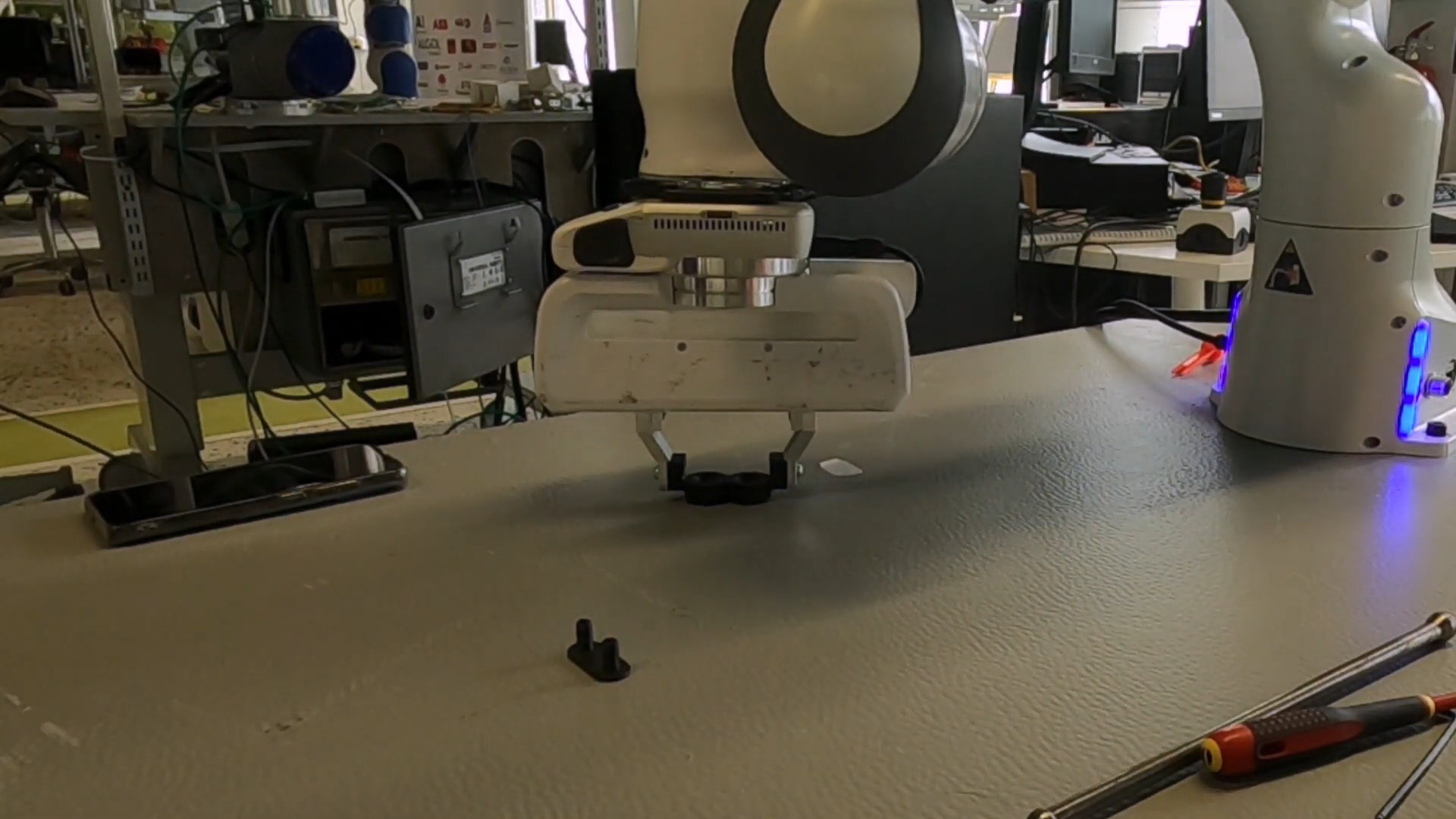} 
}
\subcaptionbox{\label{fig:Helical_6}}{%
  \includegraphics[height=0.18\linewidth]{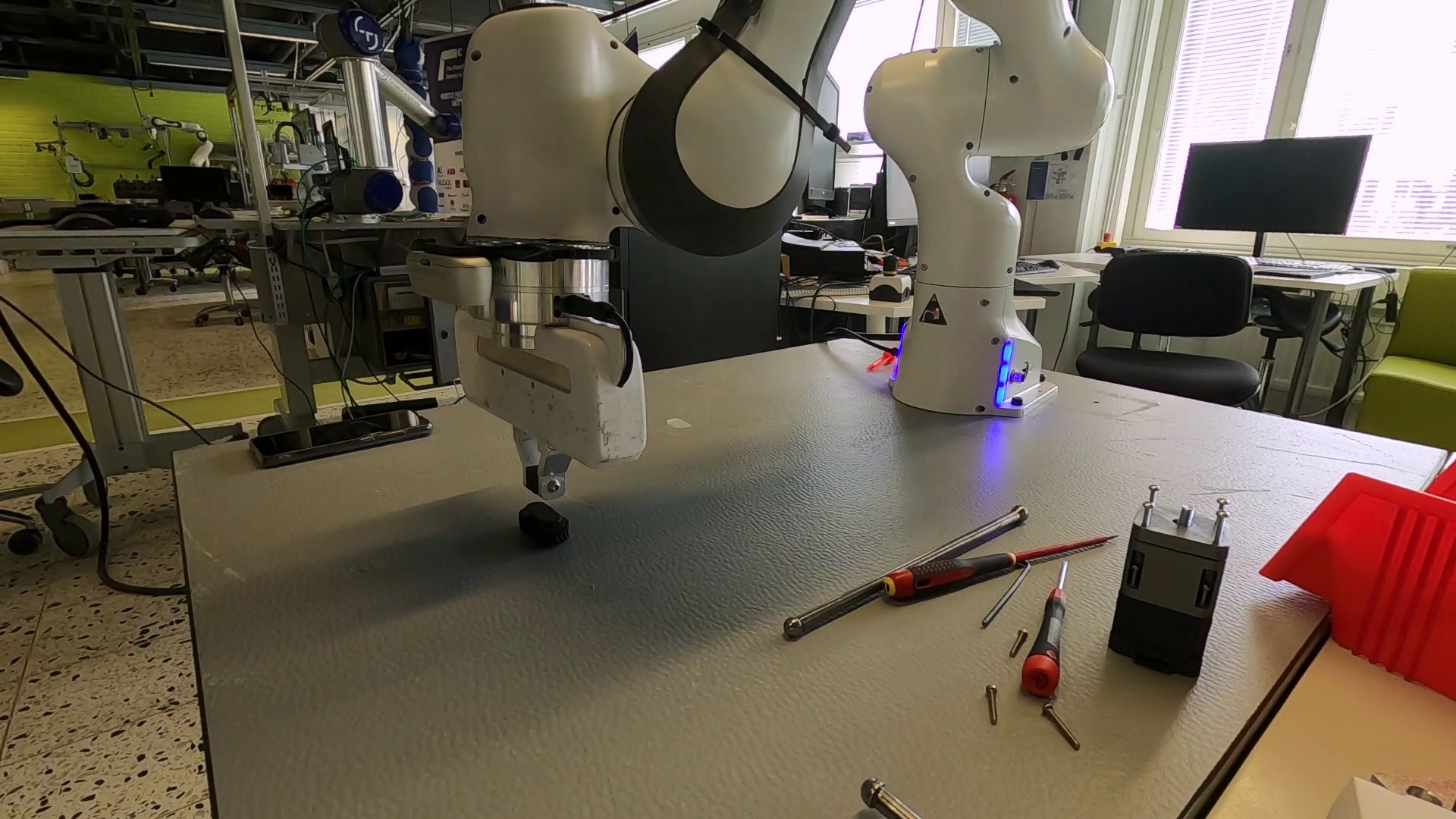} 
}
\caption{Snapshots of the verbally commanded helical gear assembly task: (a) pick back plate with {\small$\texttt{pick part}$}, (b) pick gear with {\small$\texttt{pick part}$}, (c) place gear with {\small$\texttt{place gear three}$}, (d) push gears together with {\small$\texttt{forward ninety}$}, (e) pick both gears with {\small$\texttt{close}$} and (f) place top plate with {\small$\texttt{down three}$}. 
\label{fig:helical}}

\subcaptionbox{\label{fig:planetary_1}}{%
  \includegraphics[height=0.205\linewidth]{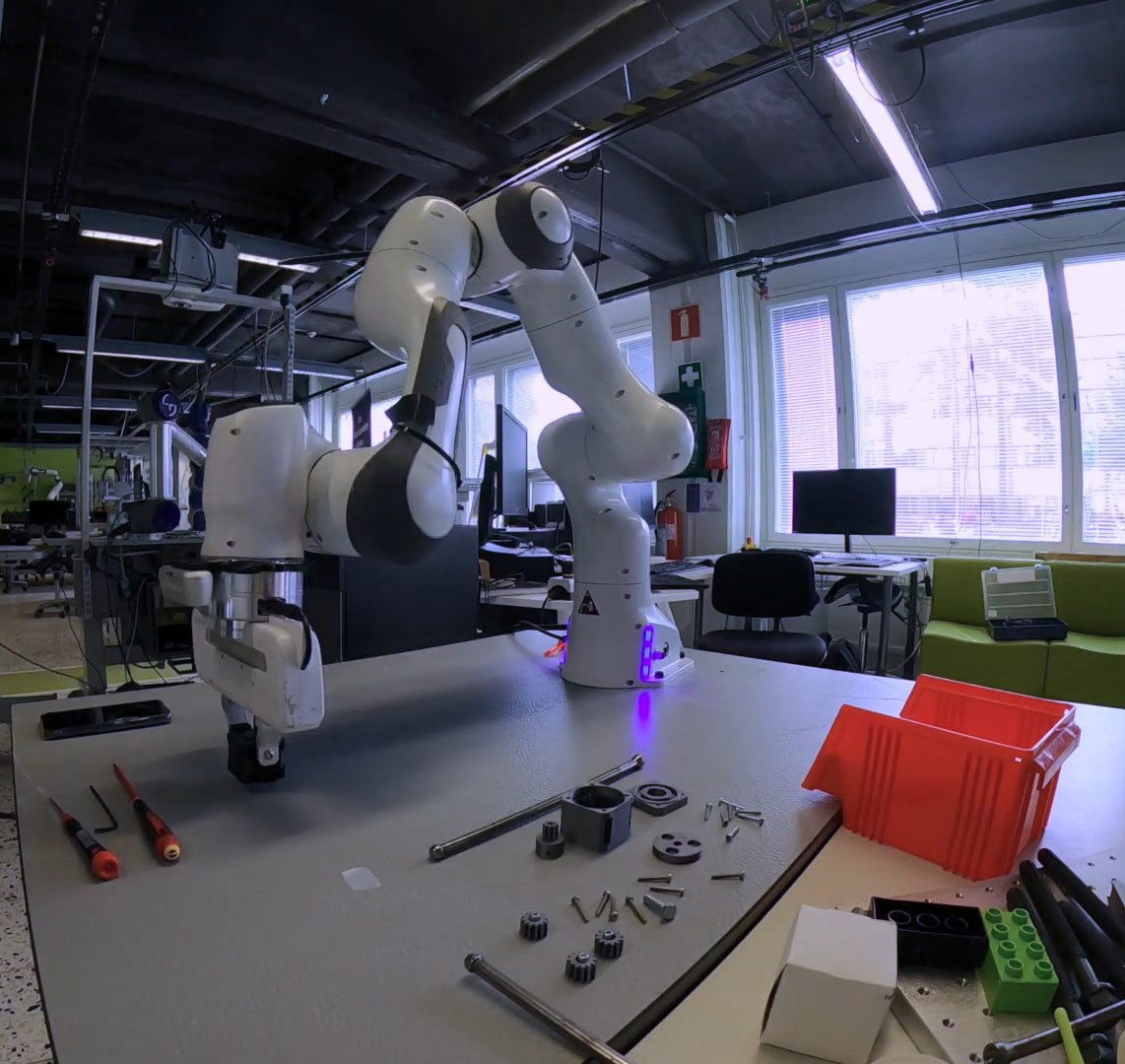} 
}\hfill
\subcaptionbox{\label{fig:planetary_2}}{%
  \includegraphics[height=0.205\linewidth]{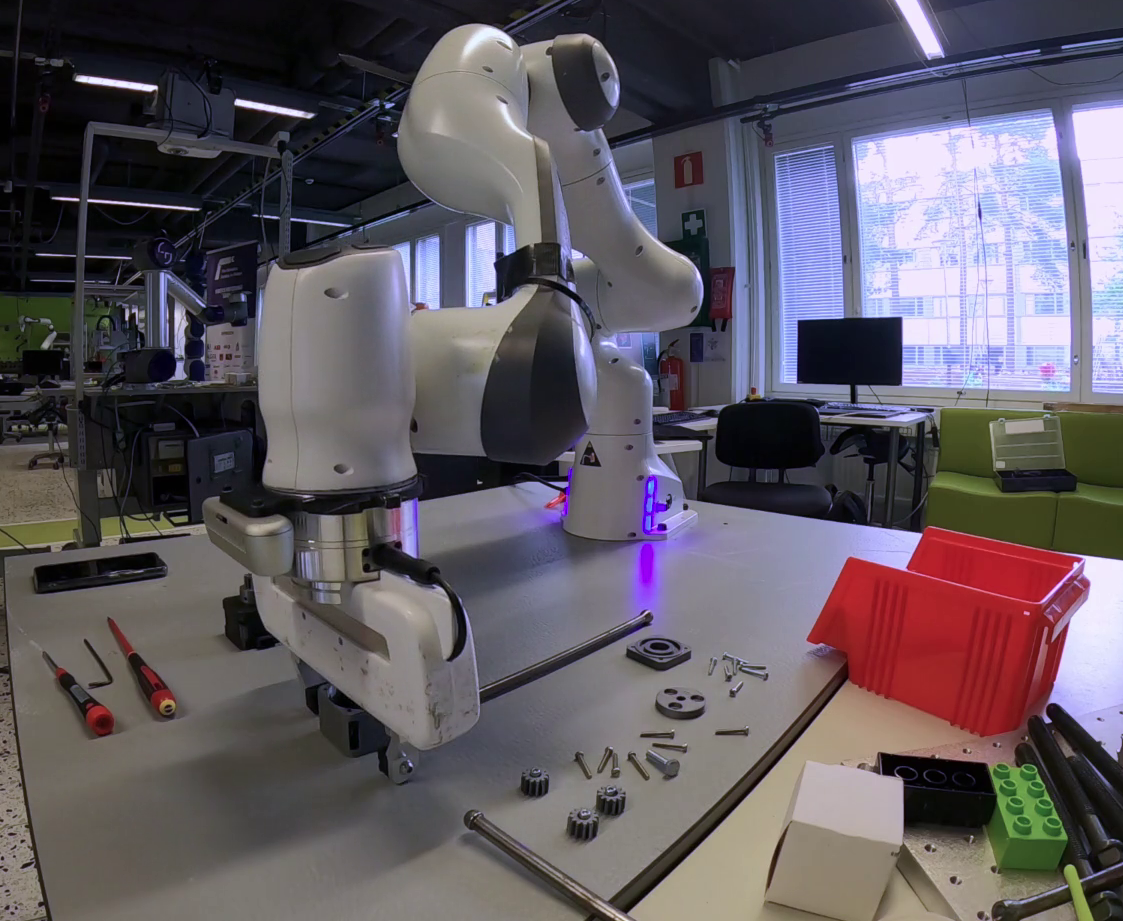} 
}\hfill
\subcaptionbox{\label{fig:planetary_3}}{%
  \includegraphics[height=0.205\linewidth]{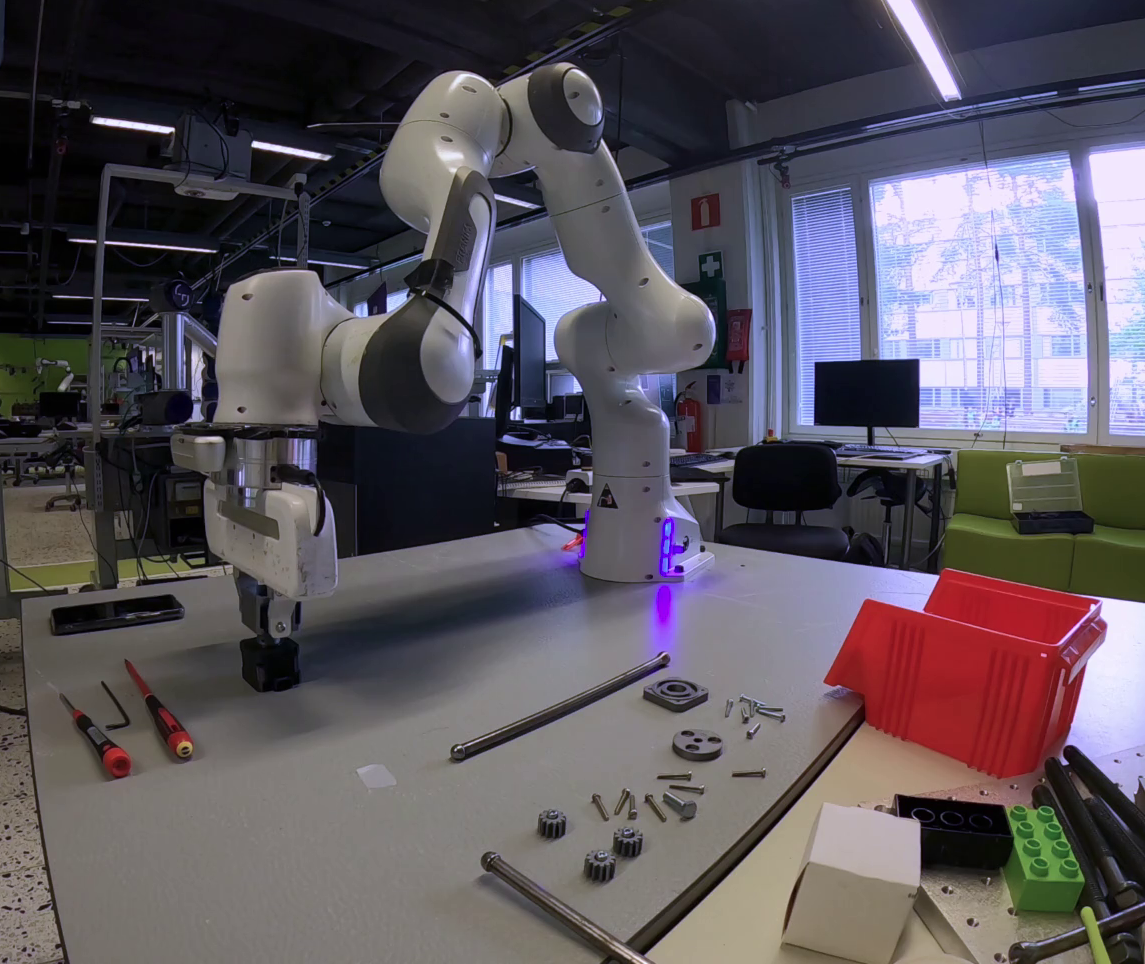} 
}\hfill
\subcaptionbox{\label{fig:planetary_4}}{%
  \includegraphics[height=0.205\linewidth]{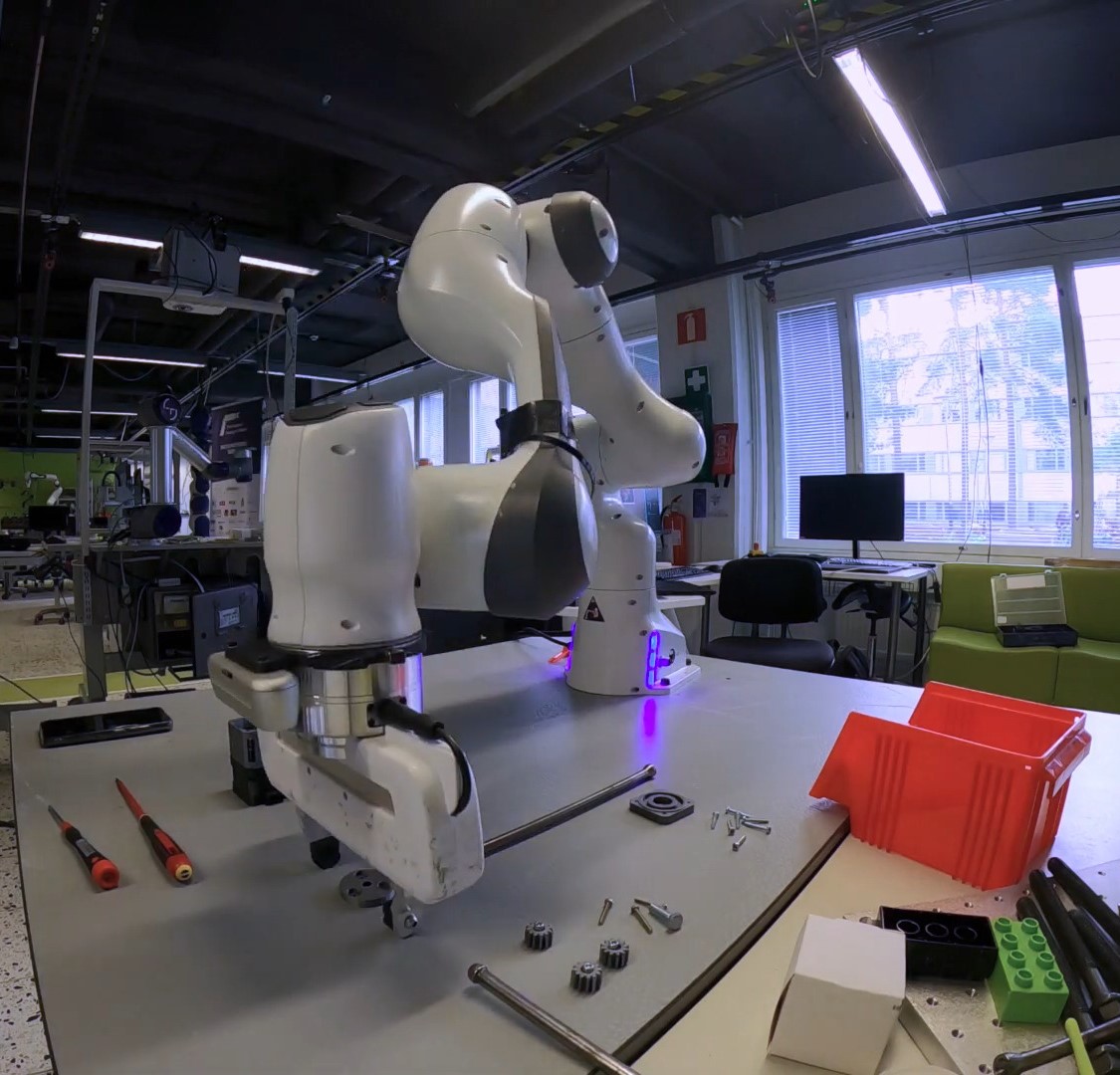} 
}
\subcaptionbox{\label{fig:planetary_5}}{%
  \includegraphics[height=0.21\linewidth]{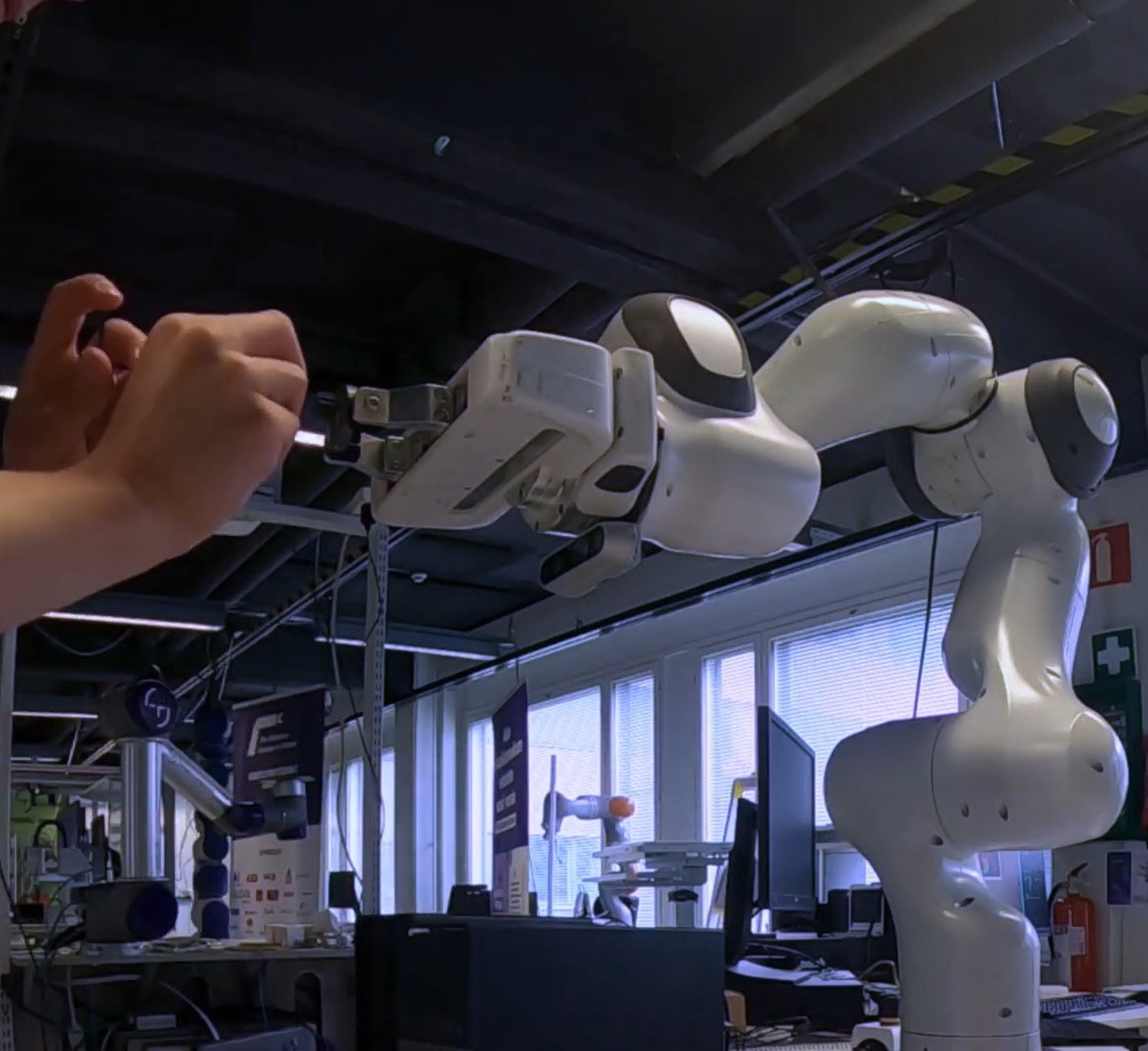} 
}\hfill
\subcaptionbox{\label{fig:planetary_6}}{%
  \includegraphics[height=0.21\linewidth]{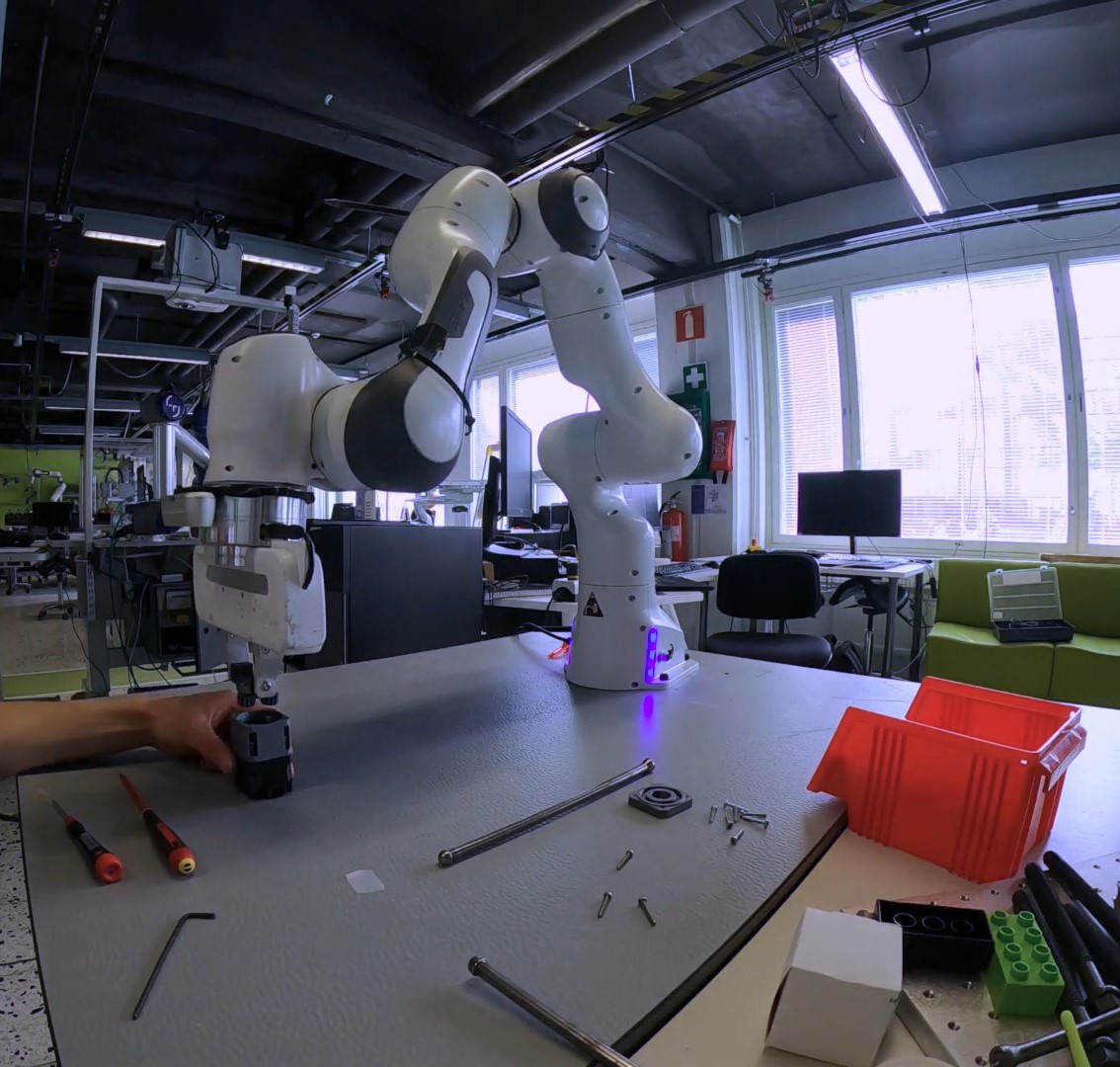} 
}\hfill
\subcaptionbox{\label{fig:planetary_7}}{%
  \includegraphics[height=0.21\linewidth]{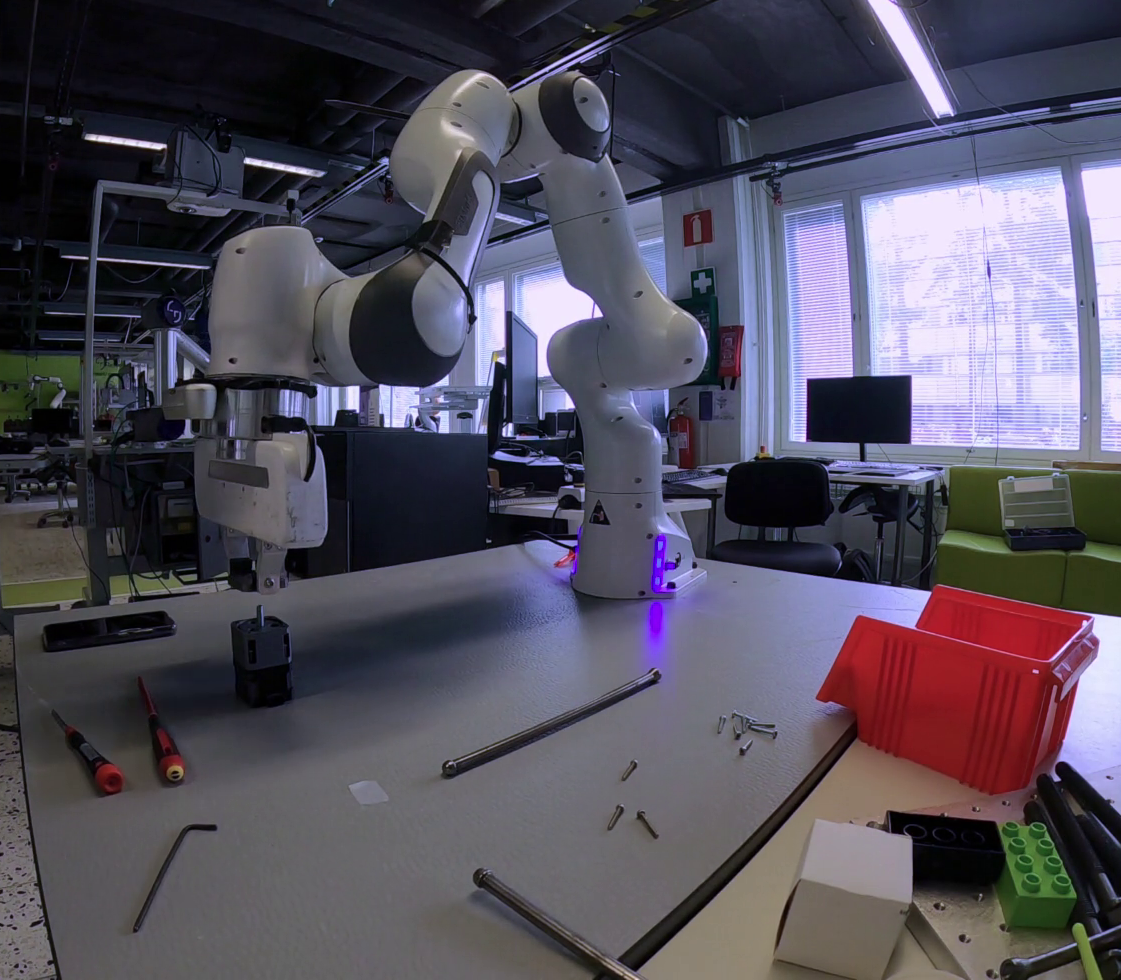} 
}\hfill
\subcaptionbox{\label{fig:planetary_8}}{%
  \includegraphics[height=0.21\linewidth]{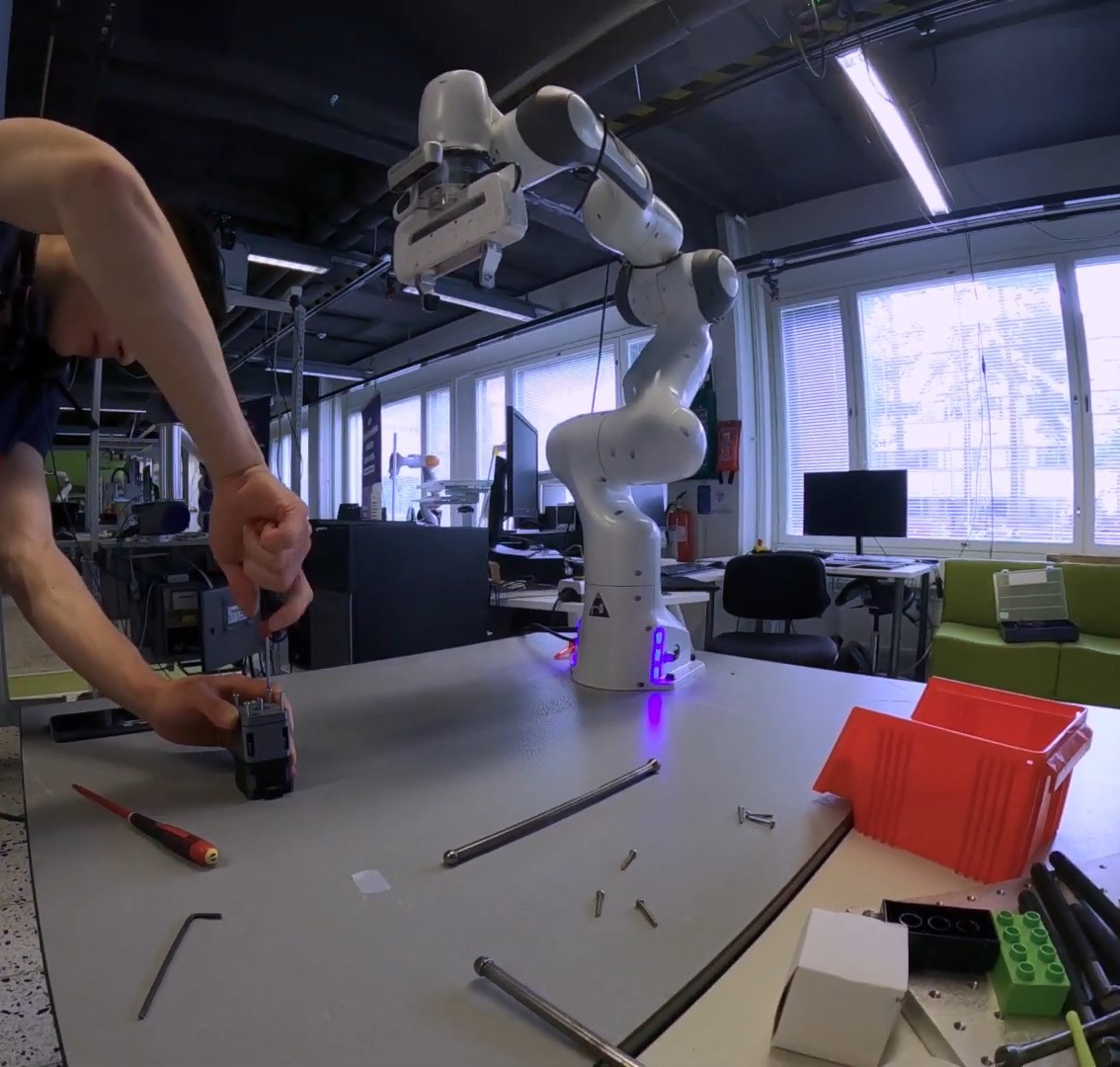} 
}
  \caption{Snapshots of the planetary gear assembly task: (a) place motor with {\small$\texttt{hold table distance thirty}$}, (b) pick housing with {\small$\texttt{pick part}$}, (c) place housing on motor with {\small$\texttt{stack table distance fifty}$}, (d) pick sun gear with {\small$\texttt{pick part}$}, (e) operator fixes planet gears on sun gear, (f) place sun gear in housing with  {\small$\texttt{hold table distance one hundred}$}, (g) place top plate on housing with {\small$\texttt{place top}$} (h) operator fixes top plate to housing.\label{fig:planetary}}
\end{figure*}

The planetary gear consists of seven major parts (housing, sun gear, three planet gears and top plate, excluding bolts, nuts and small gears) and requires nine major steps for assembly, six executed by the robot and three by a human operator. This includes five pick-and-place actions with different locations, one holding action and three human assembly actions (i.e., fixing the housing with four bolts, fixing planet gears on the sun gear and fixing the top plate with four bolts). Fig. \ref{fig:helical} depicts several snapshots of the planetary gear assembly task.

In the preparation phase of both assembly cases four poses are demonstrated to the robot, by utilizing both hand-guiding and verbal commands. In addition, in the preparation and execution phases, a high number of speech commands are used (see Table \ref{tab:results_speech_tasks}), as the assembly steps require fine adjustments before executing the assembly steps.

\begin{table*}[]
\centering
\caption{Results for speech commanded tasks. The preparation phase aims to define poses and record task sequences, by both verbal commands and hand-guiding. The execution phase entails the actual task, coordinated by the person with verbal commands. Results for both hierarchical tasks were obtained from ten trials.}
\label{tab:results_speech_tasks}
\begin{tabular}{r|ccccccc|}
\multicolumn{1}{l|}{} &
  \multicolumn{5}{c|}{\textbf{Single tasks}} &
  \multicolumn{2}{c|}{\textbf{Hierarchical tasks}} \\ \hline
\multicolumn{1}{|l|}{} &
  \multicolumn{1}{c|}{\textbf{Pick}} &
  \multicolumn{1}{c|}{\textbf{Place}} &
  \multicolumn{1}{c|}{\textbf{Pick \& Place}} &
  \multicolumn{1}{c|}{\textbf{Wipe}} &
  \multicolumn{1}{c|}{\textbf{Polish}} &
  \multicolumn{1}{c|}{\textbf{\begin{tabular}[c]{@{}c@{}}Helical gear\\ assembly\end{tabular}}} &
  \textbf{\begin{tabular}[c]{@{}c@{}}Planetary gear\\ assembly\end{tabular}} \\ \hline
\multicolumn{1}{|l|}{\textbf{Preparation phase}} &
  \multicolumn{7}{l|}{} \\ \hline
\multicolumn{1}{|r|}{Prerecorded poses} &
  \multicolumn{1}{c|}{1} &
  \multicolumn{1}{c|}{1} &
  \multicolumn{1}{c|}{3} &
  \multicolumn{1}{c|}{3} &
  \multicolumn{1}{c|}{2} &
  \multicolumn{1}{c|}{4} &
  4 \\ \hline
\multicolumn{1}{|r|}{\begin{tabular}[c]{@{}r@{}}No. of commands\\ (single, multiple)\end{tabular}} &
  \multicolumn{1}{c|}{\begin{tabular}[c]{@{}c@{}}12\\ (8/4)\end{tabular}} &
  \multicolumn{1}{c|}{\begin{tabular}[c]{@{}c@{}}11\\ (6/5)\end{tabular}} &
  \multicolumn{1}{c|}{\begin{tabular}[c]{@{}c@{}}27\\ (16/11)\end{tabular}} &
  \multicolumn{1}{c|}{\begin{tabular}[c]{@{}c@{}}22\\ (16/6)\end{tabular}} &
  \multicolumn{1}{c|}{\begin{tabular}[c]{@{}c@{}}5\\ (2/3)\end{tabular}} &
  \multicolumn{1}{c|}{\begin{tabular}[c]{@{}c@{}}44\\ (15/21)\end{tabular}} &
  \begin{tabular}[c]{@{}c@{}}34\\ (11/20)\end{tabular} \\ \hline
\multicolumn{1}{|r|}{\begin{tabular}[c]{@{}r@{}}Speech recognition succes rate\end{tabular}} &
  \multicolumn{1}{c|}{\begin{tabular}[c]{@{}c@{}}12/12\\ 100\%\end{tabular}} &
  \multicolumn{1}{c|}{\begin{tabular}[c]{@{}c@{}}11/11\\ 100\%\end{tabular}} &
  \multicolumn{1}{c|}{\begin{tabular}[c]{@{}c@{}}27/27\\ 100\%\end{tabular}} &
  \multicolumn{1}{c|}{\begin{tabular}[c]{@{}c@{}}22/22\\ 100\%\end{tabular}} &
  \multicolumn{1}{c|}{\begin{tabular}[c]{@{}c@{}}5/5\\ 100\%\end{tabular}} &
  \multicolumn{1}{c|}{\begin{tabular}[c]{@{}c@{}}36/44\\ 82\%\end{tabular}} &
  \begin{tabular}[c]{@{}c@{}}31/34\\ 91\%\end{tabular} \\ \hline
\multicolumn{1}{|r|}{Duration} &
  \multicolumn{1}{c|}{0:45} &
  \multicolumn{1}{c|}{0:30} &
  \multicolumn{1}{c|}{1:28} &
  \multicolumn{1}{c|}{1:26} &
  \multicolumn{1}{c|}{0:36} &
  \multicolumn{1}{c|}{6:12} &
  3:14 \\ \hline
\multicolumn{1}{|l|}{\textbf{Execution phase}} &
  \multicolumn{7}{c|}{} \\ \hline
\multicolumn{1}{|r|}{No. of automated steps} &
  \multicolumn{1}{c|}{1/1} &
  \multicolumn{1}{c|}{1/1} &
  \multicolumn{1}{c|}{2/2} &
  \multicolumn{1}{c|}{1/1} &
  \multicolumn{1}{c|}{1/1} &
  \multicolumn{1}{c|}{6/6} &
  6/8 \\ \hline
\multicolumn{1}{|r|}{\begin{tabular}[c]{@{}r@{}}No. of commands\\ (single/multiple)\end{tabular}} &
  \multicolumn{1}{c|}{\begin{tabular}[c]{@{}c@{}}6\\ (4/2)\end{tabular}} &
  \multicolumn{1}{c|}{\begin{tabular}[c]{@{}c@{}}6\\ (4/2)\end{tabular}} &
  \multicolumn{1}{c|}{\begin{tabular}[c]{@{}c@{}}17\\ (11/6)\end{tabular}} &
  \multicolumn{1}{c|}{\begin{tabular}[c]{@{}c@{}}6\\ (2/3)\end{tabular}} &
  \multicolumn{1}{c|}{\begin{tabular}[c]{@{}c@{}}3\\ (1/2)\end{tabular}} &
  \multicolumn{1}{c|}{\begin{tabular}[c]{@{}c@{}}55\\ (21/26)\end{tabular}} &
  \begin{tabular}[c]{@{}c@{}}65\\ (25/28)\end{tabular} \\ \hline
\multicolumn{1}{|r|}{\begin{tabular}[c]{@{}r@{}}Speech recognition succes rate\end{tabular}} &
  \multicolumn{1}{c|}{\begin{tabular}[c]{@{}c@{}}6/6\\ 100\%\end{tabular}} &
  \multicolumn{1}{c|}{\begin{tabular}[c]{@{}c@{}}6/6\\ 100\%\end{tabular}} &
  \multicolumn{1}{c|}{\begin{tabular}[c]{@{}c@{}}17/17\\ 100\%\end{tabular}} &
  \multicolumn{1}{c|}{\begin{tabular}[c]{@{}c@{}}5/6\\ 83\%\end{tabular}} &
  \multicolumn{1}{c|}{\begin{tabular}[c]{@{}c@{}}3/3\\ 100\%\end{tabular}} &
  \multicolumn{1}{c|}{\begin{tabular}[c]{@{}c@{}}46/55\\ 83\%\end{tabular}} &
  \begin{tabular}[c]{@{}c@{}}53/65\\ 82\%\end{tabular} \\ \hline
\multicolumn{1}{|r|}{Duration} &
  \multicolumn{1}{c|}{0:23} &
  \multicolumn{1}{c|}{0:22} &
  \multicolumn{1}{c|}{1:06} &
  \multicolumn{1}{c|}{2:20} &
  \multicolumn{1}{c|}{0:30} &
  \multicolumn{1}{c|}{4:18} &
  8:50 \\ \hline
\end{tabular}
\end{table*}

\subsection{Discussion}
This work demonstrated how verbal commands can instruct robot actions to achieve single and hierarchical tasks. 
The approach allows for the fast generation of robot actions and tasks, directly instructed by a person and without the need for robot programming or training a model (see Table \ref{tab:commands_vs_learning} and \cite{Shridhar2022cliport, driess2023palm}). Compared to offline programming, benefits can be especially identified when changes (e.g., pick-up location, motion sequence) need to be made during or after task execution. Speech commands allow for direct adaptation of robot motion parameters, as compared to the more traditional software practices (i.e., code compilation and evaluation). In addition, the functionalities for task concatenation enable long-term sequences of actions to be scheduled, without explicit task planners and without complex structures for task dependencies.


Natural language interaction has the benefit of expressivity \cite{marge2022spoken}, as tasks can be instructed by commands selected by the operator in any desired level of granularity. This means that accurate motions involving contact can be done more careful and under supervision of the operator (see Fig. \ref{fig:planetary_6}), leading to real-time interaction between human and robot. To ensure a responsive system, speech recognition is continuously listening for input commands and all tasks can be interrupted and overwritten by new tasks, as decided by the human operator.

Limitations of our work can be identified as well, and relate to the use of only speech for instructing commands, requiring human actions and feedback for successful robot tasks. For example, objects need to be placed on known or instructed locations and fine-grained verbal instructions are needed for delicate and contact manipulation actions. Vision as additional modality would provide a solution to these limitations, either as single perception model for object detection \cite{villani2018survey} or as visual-speech model that combines speech commands with low-level action sequences \cite{lynch2022interactive}. 

One additional limitation is that no formal planner is utilized for constructing a task plan. While this keeps the sequencing of hierarchical tasks simple, it also means that pre-conditions or dependencies between tasks are not included. Moreover, analysis on whether a (hierarchical) task sequence is suitable or optimal needs to be done by the human operator. 

Failure cases of our work were mostly caused by the limitations in automatic speech recognition (see Table \ref{tab:speech_recog_experiment} and \ref{tab:results_speech_tasks}). While current state-of-the-art language models are very robust, still disturbances occurred in case of verbal accents and words that are similar in sound (e.g., 'four' and 'for', 'tool' and 'two'). These issues were avoided by intentional clear speech pronunciations and excluding words from speech recognition that were often mis-recognized. As can be seen in the videos, few times commands had to be repeated to achieve correct speech recognition. 

Future work will investigate how Large Language Models (LLMs) can be utilized to assist in interpreting high-level verbal commands, such that more natural sentences can be instructed to the robot, instead of a predefined vocabulary. In addition, LMMs could provide additional functionalities, such as generating hierarchical tasks or complete assembly sequences automatically in the most effective and efficient manner.








\section{CONCLUSION}\label{sec:conclusion}
This work presented a framework for instructing robot action commands by human natural language. Instructions can be both fine-grained and provide the functionalities for single and hierarchical tasks, as specified by a human operator in real-time. Instructing verbal commands includes  a preparation phase were all required robot poses are demonstrated to the robot, utilizing both hand-guiding and verbal commands. In addition, hierarchical tasks can be composed to combine and concatenate robot actions. The execution phase entails the actual assembly task, coordinated by the person with verbal commands.
Experiments with several single and hierarchical assembly tasks demonstrate that verbal commands can replace traditional robot programming techniques, and provide a more expressive means to assign robot actions and enable human-robot collaboration. To provide more natural language for instructions, future work will investigate how Large Language Models can be utilized in the framework.

\section*{Acknowledgements}
Project funding was received from European Union's Horizon 2020 research and innovation programme, grant agreement no. 871252 (METRICS).
The authors thank all students in the Robotics Project Work course at Tampere University that contributed to the developments of the speech tool:  Wajeh Ahmad, Ara Jo, Omar Hassan, Tuomas Kiviö, Mikko Kulju, Waleed Rafi, Niklas Sorri and Kalle Tanninen. 

\bibliographystyle{IEEEtran}
\bibliography{IEEEabrv,refs}

\end{document}